\documentclass{article}

\usepackage{arxiv}
\usepackage[utf8]{inputenc} % allow utf-8 input
\usepackage[T1]{fontenc}    % use 8-bit T1 fonts
\usepackage{hyperref}       % hyperlinks
\usepackage{url}            % simple URL typesetting
\usepackage{booktabs}       % professional-quality tables
\usepackage{amsfonts}       % blackboard math symbols
\usepackage{nicefrac}       % compact symbols for 1/2, etc.
\usepackage{microtype}      % microtypography
\usepackage{lipsum}		    % Can be removed after putting your text content
\usepackage{graphicx}
\usepackage{doi}
\usepackage{xcolor}
\usepackage{siunitx}
\usepackage{amsmath}
\usepackage{amssymb}
\usepackage{wrapfig}
\usepackage{adjustbox}

\newcommand{\dalle}{DALL$\cdot$E-2~}
\def\sfrac#1#2{{\textstyle \frac{#1}{#2}}}

\hypersetup{
    colorlinks,
    linkcolor={red!50!black},
    citecolor={blue!50!black},
    urlcolor={blue!80!black}
}

\title{Lighting (In)consistency of Paint by Text}

\author{Hany Farid \\
	Department of Electrical Engineering and Computer Sciences \\
	School of Information \\
	University of California, Berkeley \\
	\texttt{hfarid@berkeley.edu}\\
}

\date{}

\begin{document}
\maketitle
\large

\begin{abstract}
    Whereas generative adversarial networks are capable of synthesizing highly realistic images of faces, cats, landscapes, or almost any other single category, paint-by-text synthesis engines can -- from a single text prompt -- synthesize realistic images of seemingly endless categories with arbitrary configurations and combinations. This powerful technology poses new challenges to the photo-forensic community. Motivated by the fact that paint by text is not based on explicit geometric or physical models, and the human visual system's general insensitivity to lighting inconsistencies, we provide an initial exploration of the lighting consistency of \dalle synthesized images to determine if physics-based forensic analyses will prove fruitful in detecting this new breed of synthetic media.
\end{abstract}

% keywords can be removed

% =============================================================
\section{Introduction}
\label{sec:introduction}

As OpenAI moves to expand access to their impressive \dalle paint-by-text synthesis engine\footnote{\url{https://openai.com/dall-e-2}}, concerns have been raised as to how this new technology might be misused~\cite{npr2022dalle} (see also Google's Imagen\footnote{\url{https://imagen.research.google}} and Parti\footnote{\url{https://parti.research.google}}). Paint by text synthesizes high-resolution images from a single text prompt~\cite{razavi2019generating,radford2021learning,bau2021paint,yu2022scaling}, affording control of semantic content seemingly limited only by our imagination. This latest introduction into the synthetic media arena creates new challenges for the photo-forensic community.

Paint by text, and other learning-based synthesis engines, are not based on explicit modeling of a 3-D scene, lighting, or camera. It is perhaps not surprising, therefore, that some aspects of scene geometry, including perspective geometry, are not always preserved in these synthesized images~\cite{farid2022perspective}. 

Although these synthesized images appear highly natural, the human visual system is generally insensitive to lighting inconsistencies~\cite{ostrovsky05} and -- as with perspective geometry -- may not necessarily notice implausible, unnatural, or inconsistent lighting. We provide an initial exploration of the lighting consistency of \dalle synthesized images to determine if lighting (modeled using spherical-harmonic based lighting environments~\cite{ramamoorthi2001relationship,basri2003lambertian}) are globally consistent with natural images, and if lighting is locally consistent within a synthesized image. 

This analysis reveals that while global illumination is -- with a few exceptions -- relatively consistent with natural images, lighting within an image can be highly variable. These observations may prove forensically useful in distinguishing synthesized from photographed images.

% =============================================================
\section{Photo Forensics: Lighting Analysis}
\label{sec:lightingforensics}

Under the assumption of a single distant light source (e.g.,~the sun), the appearance $I_k$ of a convex Lambertian surface of constant reflection can be modeled as $I_k = \vec{L}^T \cdot \vec{N}_k + A$, where $\vec{L}$ is the 3-D vector denoting the orientation to the light source, $\vec{N}_k$ is the 3-D surface normal corresponding to pixel $k$, and $A$ is a scalar term embodying the ambient light\footnote{This lighting model assumes the angle between the light $\vec{L}$ and surface normal $\vec{N}_k$ is between $0^\circ$ and $90^\circ$.}. Despite it simplicity, this lighting model has been used to forensically analyze images for lighting consistency~\cite{johnson2005exposing,carvalho2015exposing}.

\begin{figure}[t]
\begin{center}
\begin{tabular}{ccccc}
 & & $Y_{0,0}$ \\
 & & \includegraphics[width=1.0in]{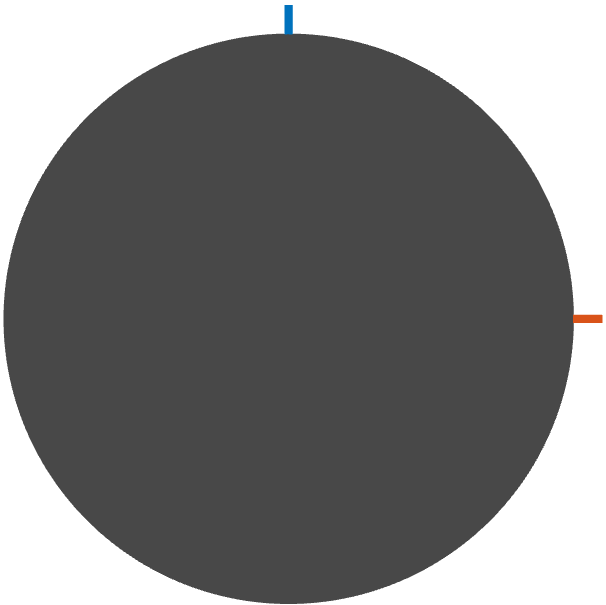} & & \\
 &   $Y_{1,-1}$ & $Y_{1,0}$ & $Y_{1,1}$ \\
 &   \includegraphics[width=1.0in]{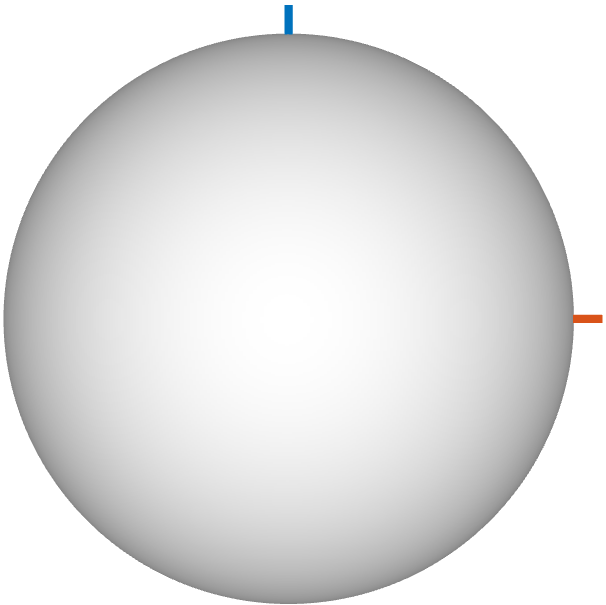} &
     \includegraphics[width=1.0in]{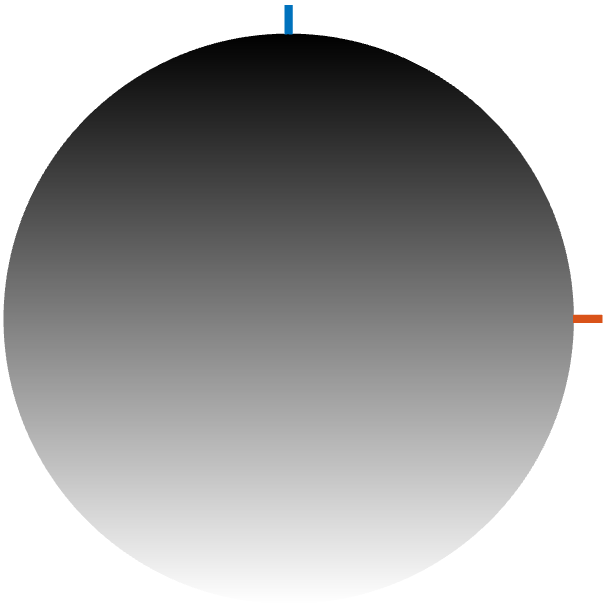} & 
     \includegraphics[width=1.0in]{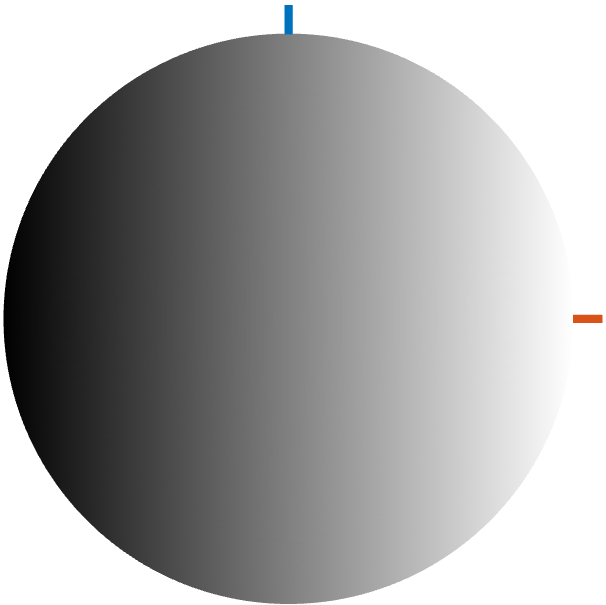} & \\
     $Y_{2,-2}$ & $Y_{2,-1}$ & $Y_{2,0}$ & $Y_{2,1}$ & $Y_{2,2}$ \\
     \includegraphics[width=1.0in]{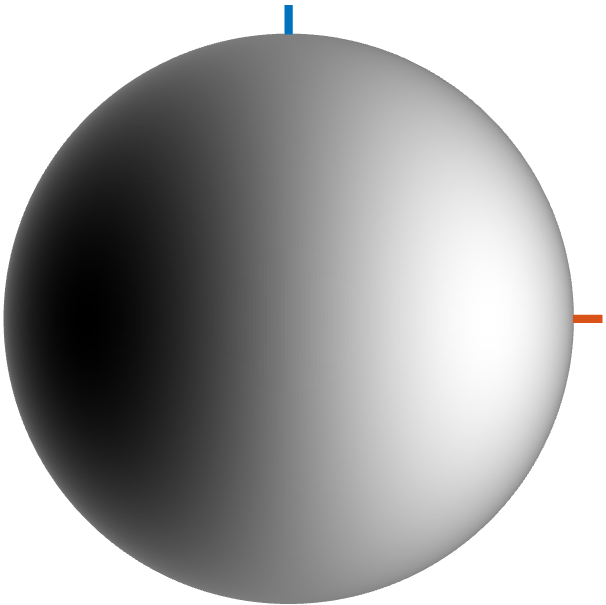} &
     \includegraphics[width=1.0in]{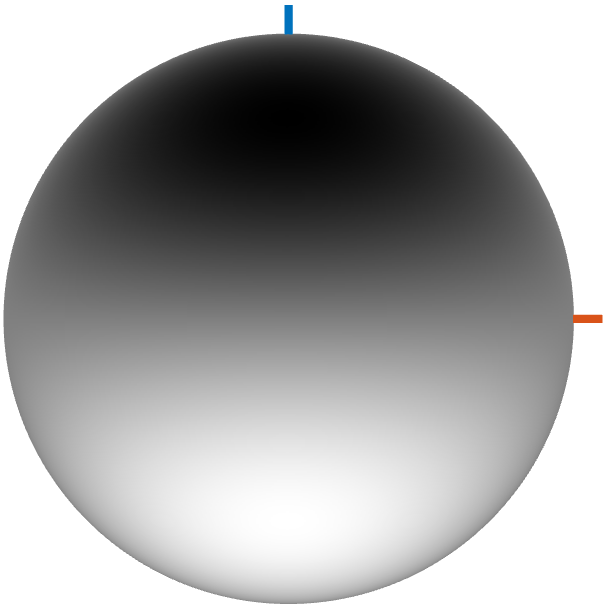} & 
     \includegraphics[width=1.0in]{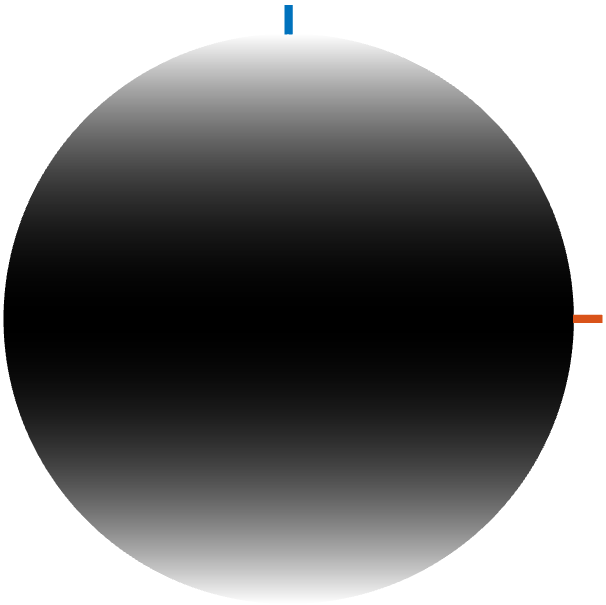} &
     \includegraphics[width=1.0in]{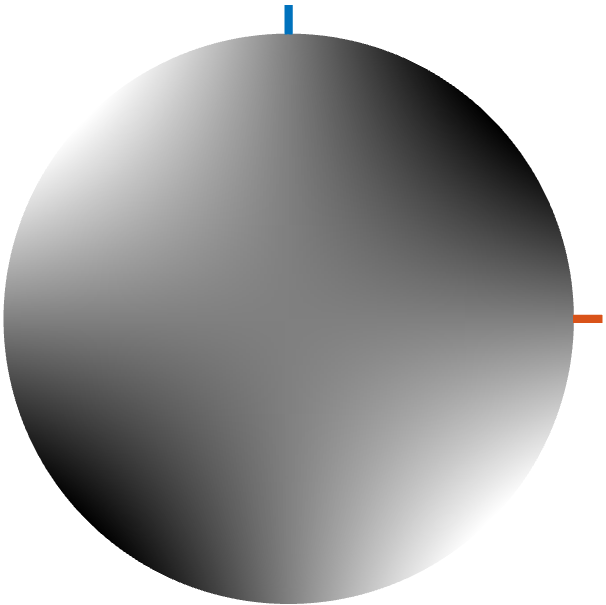} &
     \includegraphics[width=1.0in]{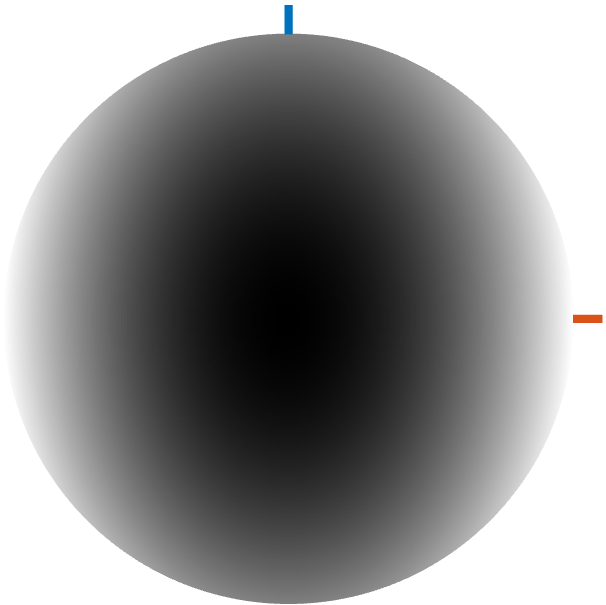}
 \end{tabular}
\end{center}
\caption{The first three orders of spherical harmonics as functions on the sphere. Shown from top to bottom are the zeroth-order spherical harmonic, $Y_{0,0}(\cdot)$; the three first-order spherical harmonics, $Y_{1,*}(\cdot)$; and the five second-order spherical harmonics, $Y_{2,*}(\cdot)$. The red and blue axis corresponds to the positive $x$- and $z$-axis (corresponding to the camera optical axis), and the positive $y$-axis is facing into the page.}
\label{fig:harmonics}
\end{figure}

This lighting model can be extended to accurately capture more complex lighting environments, with a number of arbitrarily placed light sources. As originally described in~\cite{ramamoorthi2001relationship,basri2003lambertian}, and forensically employed in~\cite{johnson2007exposing,riess2010scene,kee2010exposing,de2013exposing}, under the assumption of distant lighting, the appearance of a convex Lambertian surface of constant reflectance can be modelled with a spherical Fourier basis. Specifically, a point on a surface with 3-D surface normal $\vec{N}_k$, is imaged to pixel location $k$ with pixel intensity given by:
\begin{eqnarray}
\hspace{-0.5cm}
I_k    & = & l_{0,0} \pi Y_{0,0}(\vec{N}_k) + \nonumber \\
       &  & l_{1,-1} \sfrac{2\pi}{3} Y_{1,-1}(\vec{N}_k) +
             l_{1,0}  \sfrac{2\pi}{3} Y_{1,0}(\vec{N}_k) +
             l_{1,1}  \sfrac{2\pi}{3} Y_{1,1}(\vec{N}_k) + \nonumber \\
       &  & l_{2,-2} \sfrac{\pi}{4} Y_{2,-2}(\vec{N}_k) + 
             l_{2,-1} \sfrac{\pi}{4} Y_{2,-1}(\vec{N}_k) +
             l_{2,0}  \sfrac{\pi}{4} Y_{2,0}(\vec{N}_k) +
             l_{2,1}  \sfrac{\pi}{4} Y_{2,1}(\vec{N}_k) + l_{2,2} \sfrac{\pi}{4} Y_{2,2}(\vec{N}_k),
\label{eqn:image-intensity}           
\end{eqnarray}
where, the spherical harmonics $Y_{*,*}(\cdot)$ -- embodying the lighting environment -- are parameterized as a function of the 3-D surface normal $\vec{N}_k=\begin{pmatrix}x_k & y_k & z_k\end{pmatrix}$:
\begin{equation}
\begin{array}{r@{\hspace{1mm}}c@{\hspace{1mm}}l@{\hspace*{5mm}}r@{\hspace{1mm}}c@{\hspace{1mm}}l@{\hspace*{5mm}}r@{\hspace{1mm}}c@{\hspace{1mm}}l}
Y_{0,0} (\vec{N}_k)  & = & \frac{1}{\sqrt{4\pi}} &
Y_{1,-1}(\vec{N}_k)  & = & \sqrt{\frac{3}{4\pi}} y_k &
Y_{1,0} (\vec{N}_k)  & = & \sqrt{\frac{3}{4\pi}} z_k \\
Y_{1,1} (\vec{N}_k)  & = & \sqrt{\frac{3}{4\pi}} x_k &
Y_{2,-2}(\vec{N}_k)  & = & 3\sqrt{\frac{5}{12\pi}}x_ky_k &
Y_{2,-1}(\vec{N}_k)  & = & 3\sqrt{\frac{5}{12\pi}}y_kz_k \\
Y_{2,0} (\vec{N}_k)  & = & \frac{1}{2} \sqrt{\frac{5}{4\pi}} (3z_k^2 - 1) &
Y_{2,1} (\vec{N}_k)  & = & 3\sqrt{\frac{5}{12\pi}}x_kz_k &
Y_{2,2} (\vec{N}_k)  & = & \frac{3}{2} \sqrt{\frac{5}{12\pi}} (x_k^2 - y_k^2).
\end{array}
\label{eqn:spherical-harmonics}
\end{equation}

As shown in Figure~\ref{fig:harmonics}, the zeroth-order spherical harmonic, $Y_{0,0}(\cdot)$, embodies the ambient lighting, the first-order spherical harmonics, $Y_{1,*}(\cdot)$, embody directional lighting (top/down, front/back, and left/right), and the second order spherical harmonics, $Y_{2,*}(\cdot)$, embody higher-order, directional lighting effects.

Note that the expression of pixel intensity, Equation~(\ref{eqn:image-intensity}), is linear in the nine lighting environment coefficients, $l_{0,0}$ to $l_{2,2}$. Given 3-D surface normals at $p \geq 9$ points, the lighting environment coefficients can be estimated as the least-squares solution to the following system of linear equations:
\begin{eqnarray}
\begin{pmatrix}
\pi Y_{0,0}(\vec{N}_1) & \frac{2\pi}{3} Y_{1,-1}(\vec{N}_1) & \ldots & \frac{\pi}{4} Y_{2,2}(\vec{N}_1) \cr
\pi Y_{0,0}(\vec{N}_2) & \frac{2\pi}{3} Y_{1,-1}(\vec{N}_2) & \ldots & \frac{\pi}{4} Y_{2,2}(\vec{N}_2) \cr
\vdots & \vdots & \ddots & \vdots \cr 
\pi Y_{0,0}(\vec{N}_p) & \frac{2\pi}{3} Y_{1,-1}(\vec{N}_p) & \ldots & \frac{\pi}{4} Y_{2,2}(\vec{N}_p) \cr
\end{pmatrix}
\begin{pmatrix} 
l_{0,0} \cr
l_{1,-1} \cr
\vdots \cr
l_{2,2}
\end{pmatrix} 
& = & 
\begin{pmatrix}
I_1 \cr
I_2 \cr
\vdots \cr
I_p \cr
\end{pmatrix} \nonumber \\
A\vec{l} & = & \vec{b},
\label{eqn:light-env-twls}
\end{eqnarray}
where $A$ is the $p \times 9$ matrix containing the sampled spherical harmonics, $\vec{l}$ is the $9 \times 1$ vector of unknown lighting environment coefficients, and $\vec{b}$ is the $p \times 1$ vector of intensities at $p$ pixel locations. The least-squares solution to this system is $\vec{l} = \left( A^T A \right)^{-1} A^T \vec{b}$. This estimation can be performed for each of three color channels in an RGB image.

\begin{figure}[t]
\begin{center}
\begin{tabular}{cc}
    (a) & (b) \\
     \includegraphics[width=0.45\linewidth]{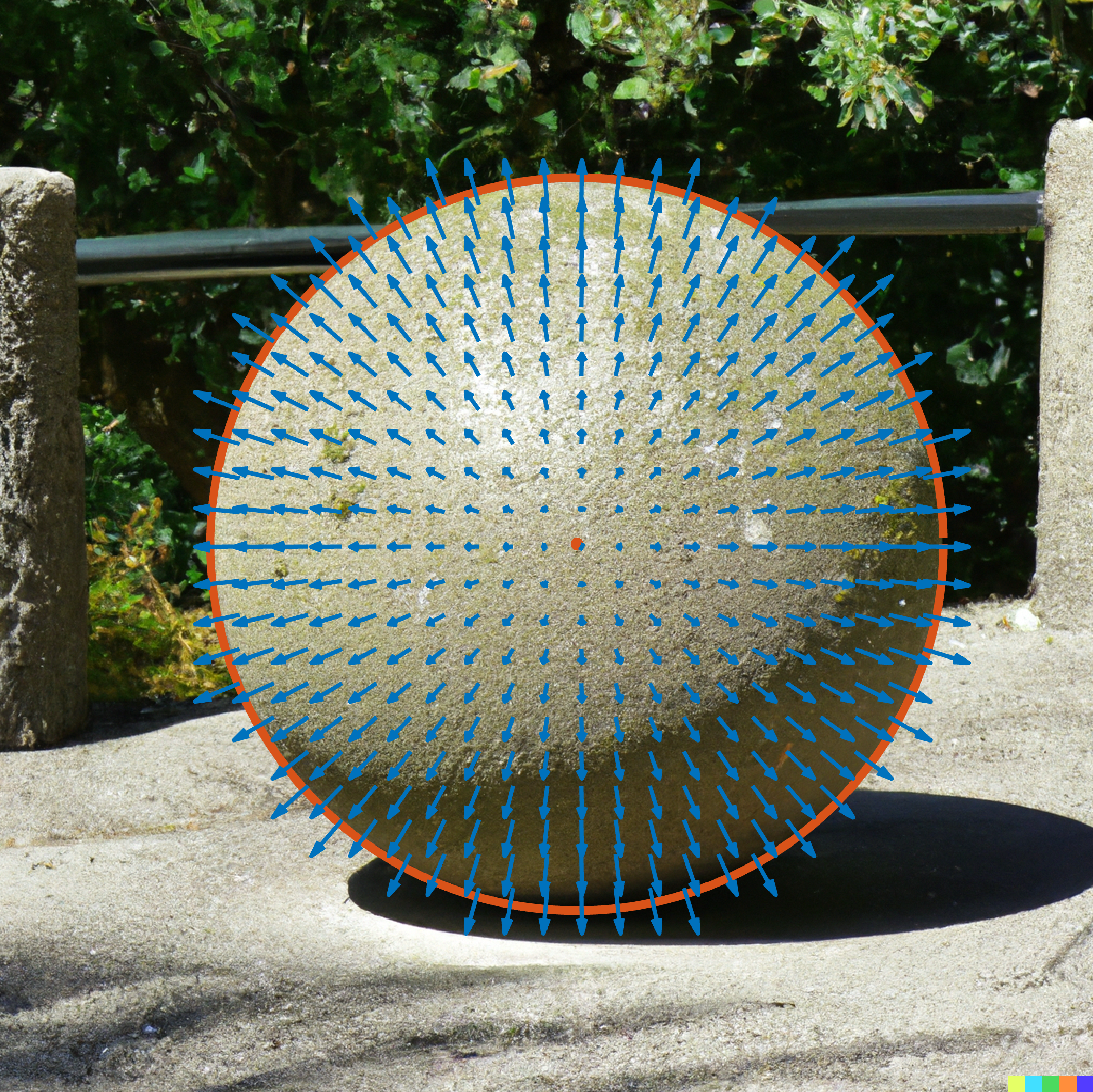} & 
     \fbox{\includegraphics[width=0.45\linewidth]{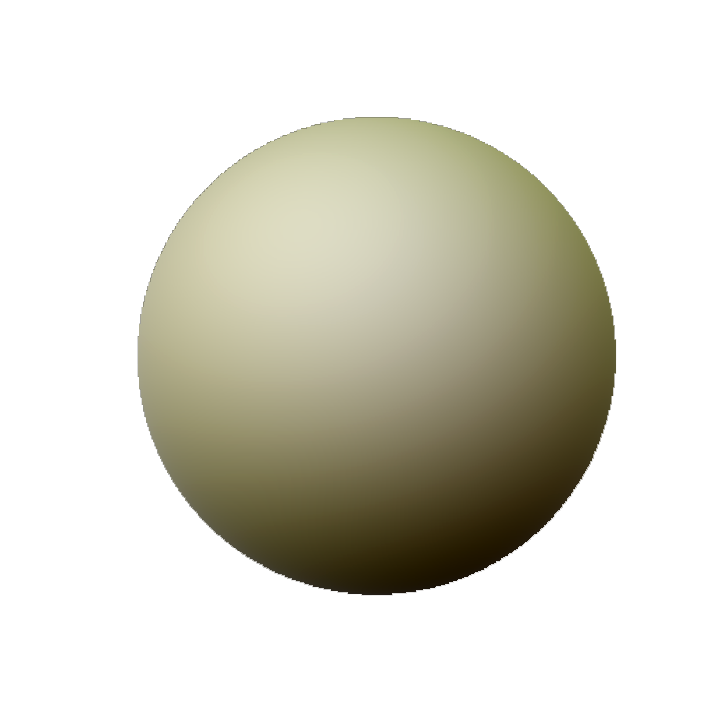}}  
\end{tabular}
\end{center}
\caption{Estimating 3-D lighting environment from a sphere: (a) a \dalle synthesized image of a concrete garden sphere overlaid with a fitted circle (red) and estimated 3-D surface normals (blue); and (b) a rendered sphere illuminated with the estimated lighting environment.}
\label{fig:garden-sphere}
\end{figure}

While the accurate estimation of 3-D surface normals may not always be possible from a single image, basic geometric shapes like a sphere afford a simple way to extract 3-D surface normals. A sphere photographed from any orientation will be imaged as an ellipse; in images recorded with a narrow field of view, these ellipses will be well approximated by a circle, particularly near the center of the image. We will assume that the image of a sphere is circular, and therefore the 3-D surface normal at any point on the sphere can be determined directly from the pixel coordinate $(x_k,y_k)$ as:
\begin{eqnarray}
\vec{N}_k & = & \begin{pmatrix} x_k-c_x & y_k-c_y & \sqrt{r^2 - (x_k-c_x)^2 + (y_k-c_y)^2} \end{pmatrix},
\label{eqn:3d-surface-normal}
\end{eqnarray}
where $r$ and $(c_x,c_y)$ are the radius and center of the image-based circular projection of a sphere, and $\vec{N}_k$ is scaled to unit length: $\vec{N}_k / \| \vec{N}_k \|$.

Shown in Figure~\ref{fig:garden-sphere}(a) is an image of a concrete garden sphere overlaid with a fitted circle (see below) and a subset of the estimated 3-D surface normals. Shown in panel (b) of this figure is a sphere rendered with the estimated lighting environment, where we can see that the overall color and positioning of the lighting has been captured.

An EM-based approach is used to fit a circle (center and radius) to the image of a sphere, Figure~\ref{fig:garden-sphere}(a). An RGB image is first converted to grayscale and histogram equalized (to boost edge contrast) followed by a standard gradient-based edge detection and an intensity threshold to yield a binary-valued image. The resulting binary image highlights the salient edges, including -- but not exclusively -- the circle's boundary. An expectation/maximization (EM) algorithm~\cite{dempster1977maximum} is employed to iteratively segment the edge pixels into those belonging to the circle's boundary, and to estimate the circle center and radius. 

In the E-step, the residual error between each edge pixel $k$ is computed as the shortest distance between the pixel at location $(x_k,y_k)$ and the current estimate of the circle with center $(c_x,c_y)$ and radius ($r$):
\begin{eqnarray}
    \delta_k & = & | (x_k-c_x)^2 + (y_k-c_y)^2 - r^2 |.
\label{eqn:circle-shortest-dist}
\end{eqnarray}
The probability that any pixel $k$, with residual error $\delta_k$, is associated with the circle's boundary is computed as:
\begin{eqnarray}
    w_k & = & \frac{e^{-\delta_k^2/2\sigma^2}}{e^{-\delta_k^2/2\sigma^2} + \epsilon},
\label{eqn:estep-probability}
\end{eqnarray}
where $\sigma$ is the variance of the underlying normal distribution, and $\epsilon$ is the uniform probability that pixel $k$ is not associated with the circle's boundary (i.e.,~pixel $k$ belongs to an outlier model).

\begin{figure}[t]
\begin{center}
\begin{tabular}{c@{\hspace{0.15cm}}c@{\hspace{0.15cm}}c@{\hspace{0.15cm}}c}
     \includegraphics[width=0.225\linewidth]{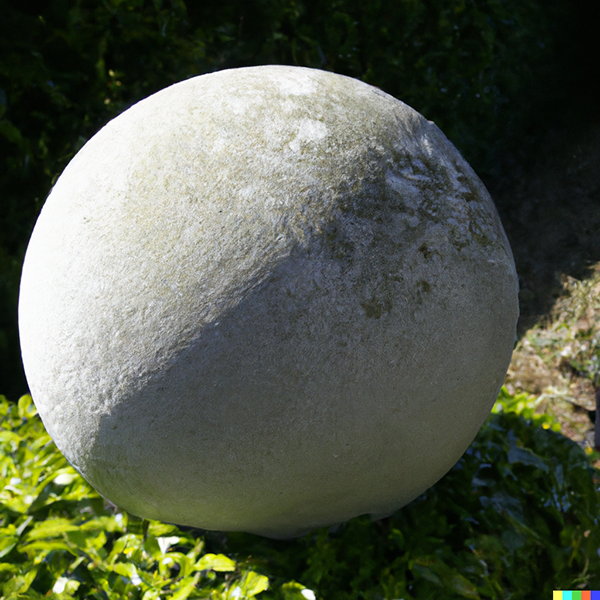} & 
     \includegraphics[width=0.225\linewidth]{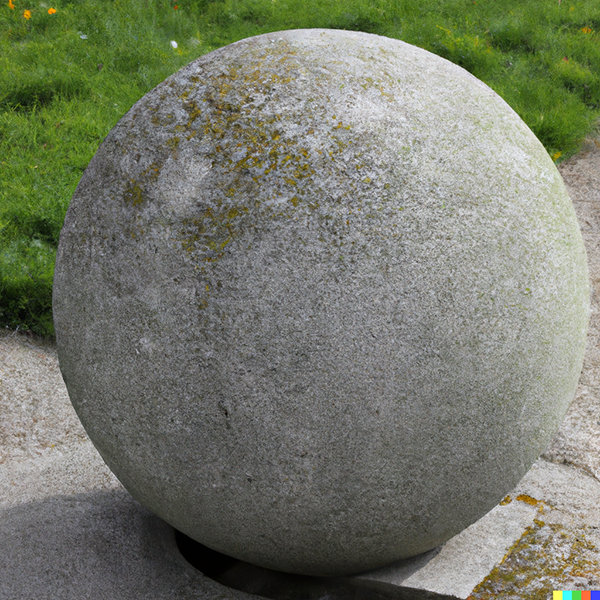} & 
     \includegraphics[width=0.225\linewidth]{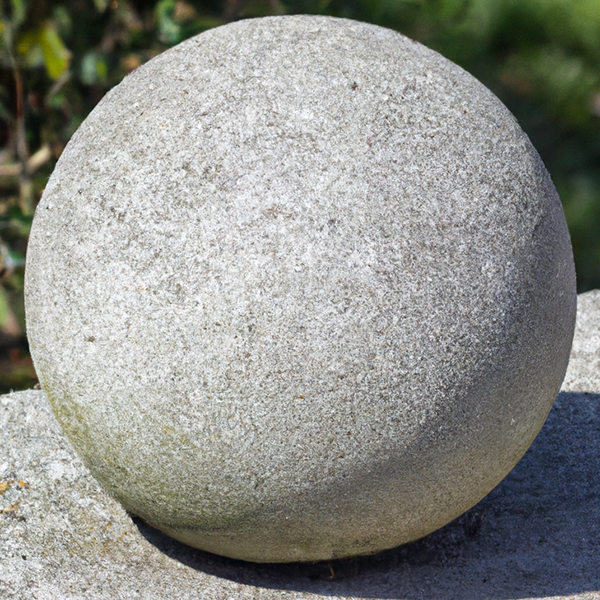} & 
     \includegraphics[width=0.225\linewidth]{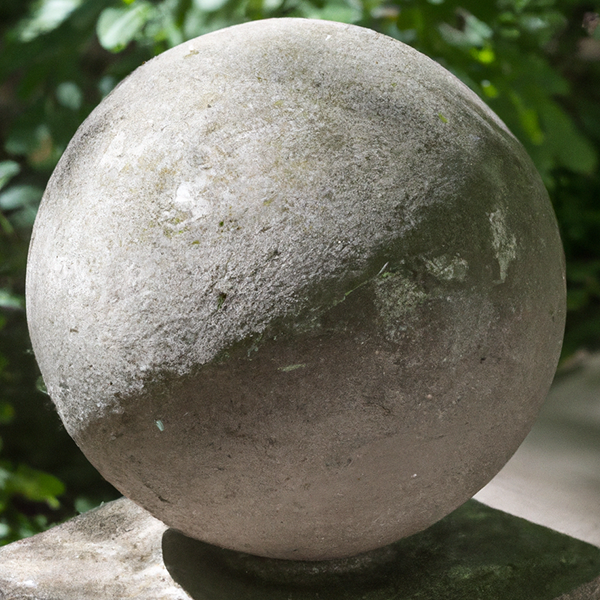} \\ 
     \includegraphics[width=0.225\linewidth]{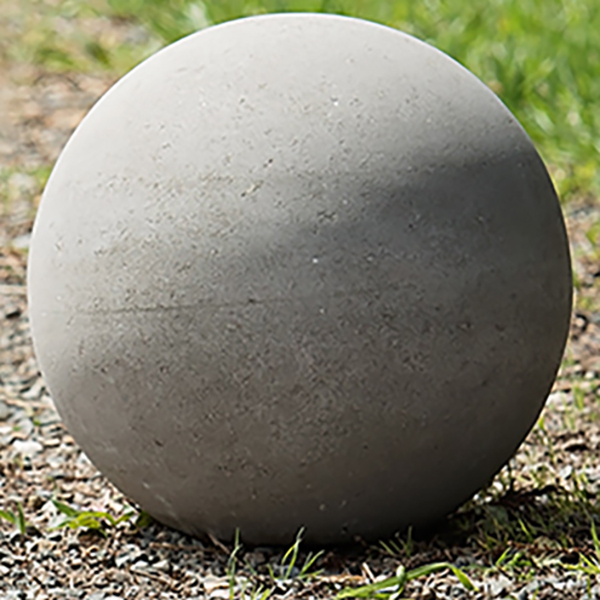} & 
     \includegraphics[width=0.225\linewidth]{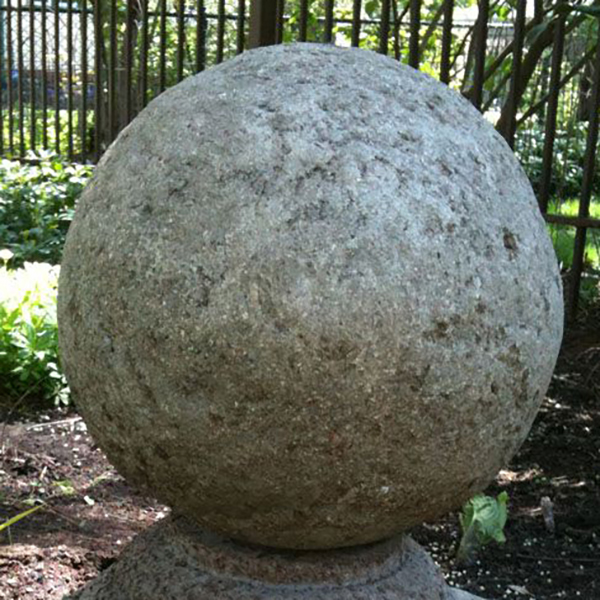} & 
     \includegraphics[width=0.225\linewidth]{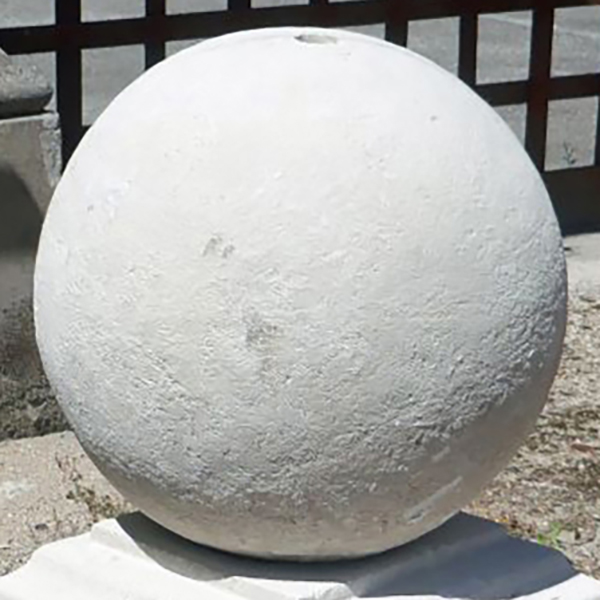} & 
     \includegraphics[width=0.225\linewidth]{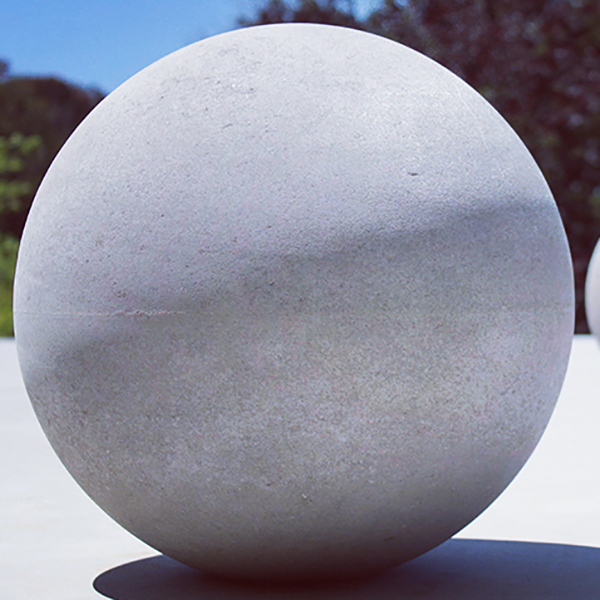}     
\end{tabular}
\end{center}
\caption{Representative examples of garden spheres used to evaluate the consistency of global lighting environments between synthesized images (top row) and photographed images (bottom row).}
\label{fig:garden-spheres-examples}
\end{figure}

In the M-step, a weighted total least-squares estimation~\cite{gander1994least} is used to re-estimate the circle's center ($c_x,c_y$) and radius ($r$) as the minimal eigenvalue-eigenvector $(\vec{v}$) of $M^TW^TWM$, where:
\begin{eqnarray}
    M ~=~ \begin{pmatrix} 
        x_1^2+y_1^2 & x_1 & y_1 & 1 \cr 
        x_2^2+y_2^2 & x_2 & y_2 & 1 \cr 
        \vdots & \vdots & \vdots & \vdots \cr 
        x_n^2+y_n^2 & x_n & y_n & 1
        \end{pmatrix} 
    \qquad \mbox{and} \qquad
    W ~=~ \begin{pmatrix}
        w_1 & 0   & \ldots & 0 \cr
        0   & w_2 & \ldots & 0 \cr
        \vdots    & \vdots & \ddots & \vdots \cr
        0   & 0   & \ldots & w_n,
        \end{pmatrix}
\label{eqn:mstep-wls}
\end{eqnarray}
where $n$ is the total number of identified edge pixels, and $w_k$ is the probability that pixel $k$ is associated with the circle's boundary, as computed in the E-step, Equation~(\ref{eqn:estep-probability}). The circle's center and radius are extracted from the solution $\vec{v}$ as follows:
\begin{eqnarray}
    c_x ~=~ -\frac{v_2}{2v_1}  \qquad 
    c_y ~=~ -\frac{v_3}{2v_1} \qquad
    r   ~=~ \sqrt \frac{v_2^2 + v_3^2}{4v_1^2} - \frac{v_4}{v_1},
\end{eqnarray}
where $v_i$ denotes the $i^{th}$ component of the 4-D eigenvector $\vec{v}$.

The E- and M-steps are iteratively evaluated until subsequent estimates of the circle center and radius are each less than a threshold of $0.5$ pixels. On each iteration, the variance $\sigma$ in Equation~(\ref{eqn:estep-probability}) is updated to $\sum_{k=1}^{n} w_kr_k^2 / \sum_{k=1}^{n} w_k$. The EM estimation is bootstrapped by manually annotating the imaged sphere's approximate center and radius.

% =============================================================
\section{Paint by Text: Lighting Analysis}

A total of $50$ images were synthesized using \dalle with the text prompt ``a photo of a concrete sphere in a garden.'' In most cases, the resulting image was semantically consistent with the prompt. The synthesized images were manually curated to remove any images that did not show the sphere in its entirety (due to clipping or occlusion) or was not relatively uniform in appearance. A second set of $50$ images were downloaded following a web-based image search with the same text prompt ``a concrete sphere in a garden.'' The same curation was applied to the surfaced images, along with a resolution constraint of at least $1024 \times 1024$ pixels -- the same resolution as the synthesized images. The context of these images -- primarily garden-supply stores -- made it highly likely the images were photographic and not computer-generated or synthesized. 

\begin{figure}[t]
\begin{center}
    \begin{tabular}{c@{\hspace{0.25cm}}|@{\hspace{0.25cm}}c}
        \begin{adjustbox}{valign=c}
            \begin{tabular}{@{}c@{}}
              photographed \\
              \includegraphics[width=0.25\linewidth]{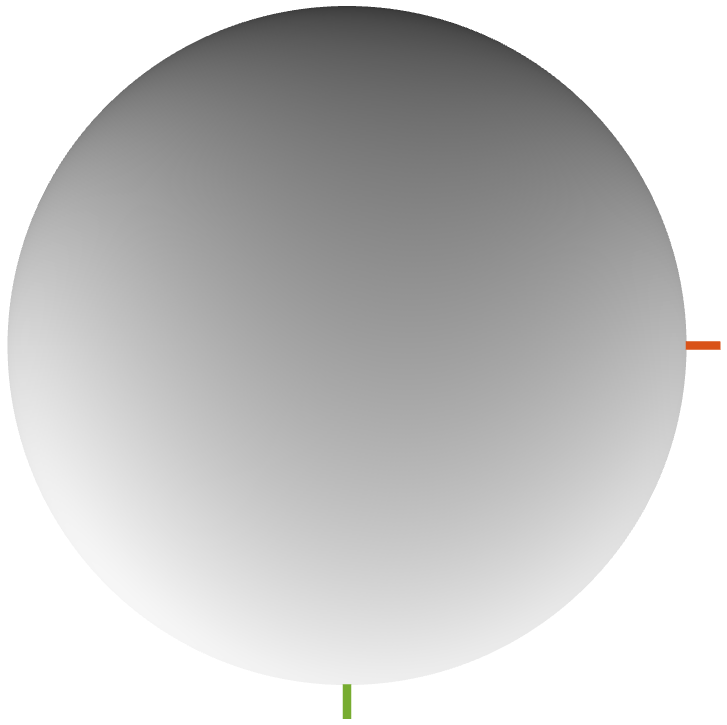} \\[2ex]
              synthesized \\ 
              \includegraphics[width=0.25\linewidth]{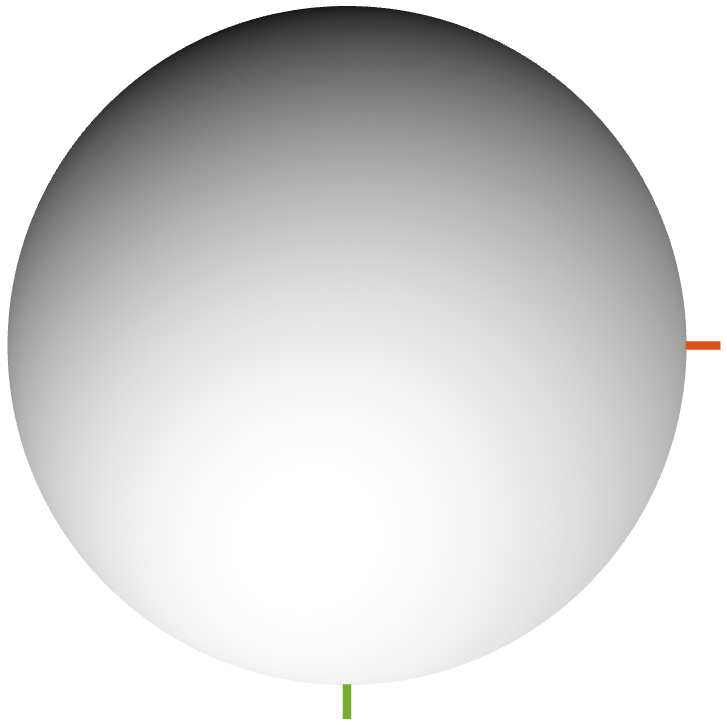}
            \end{tabular}
        \end{adjustbox}
        &
        \begin{adjustbox}{valign=c}
          \includegraphics[width=0.7\linewidth]{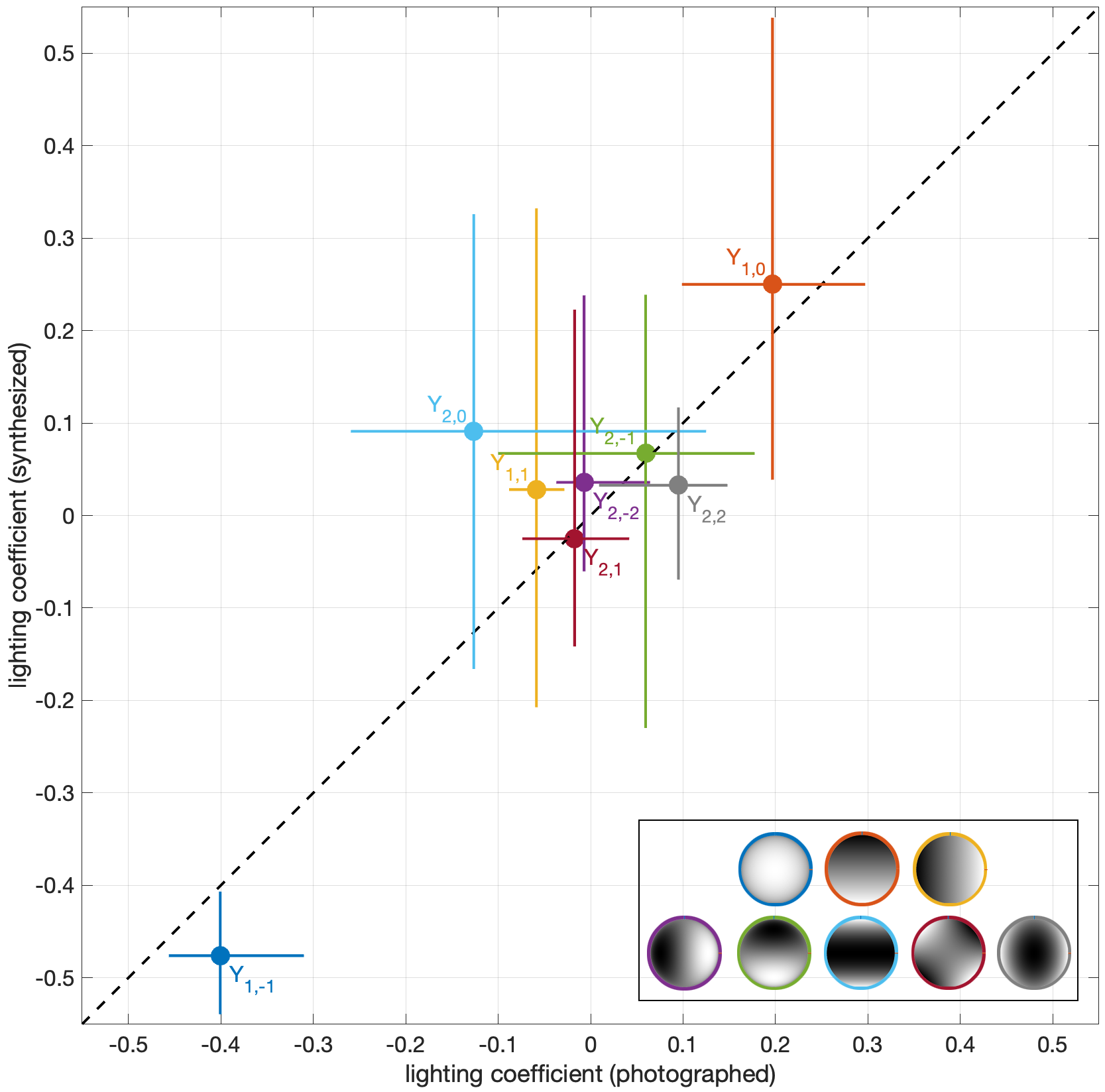}
        \end{adjustbox}
    \end{tabular}
\end{center}
\caption{Shown on the right are the environmental lighting coefficients for photographed (horizontal axis) and synthesized (vertical axis) images, with a correlation of $R^2=0.80$. Each data point corresponds to the median lighting coefficient ($50\%$ quantile) and the horizontal and vertical bars correspond to the $35\%$ and $65\%$ quantile. The first-order, $Y_{1,*}$, and second-order, $Y_{2,*}$, coefficients are normalized by the zeroth-order coefficient $Y_{0,0}$. Shown on the left is a sphere rendered with the median photographic (top) and synthesized (bottom) lighting environment (these spheres are displayed on a shared intensity range and the red and green axes correspond to the $x$- and $y$-axis and the $z$-axis (camera optical axis) is facing into the page -- note these spheres are rendered with a different viewpoint than the legend to highlight the differences between them).}
\label{fig:lighting-env-results}
\end{figure}

All $100$ images were manually cropped loosely around the sphere and uniformly resized to $600 \times 600$ pixels. Shown in Figure~\ref{fig:garden-spheres-examples} are four representative examples of these synthesized (top row) and photographed (bottom row) images. 

After fitting a circle to the image of the sphere, the 3-D surface normals are estimated, Figure~\ref{fig:garden-sphere}; given the relatively high image resolution, the normals are sampled at every other pixel. The image is then spatially filtered with a $9$-tap median filter to minimize the impact of non-uniformities in the sphere's reflectance. The lighting environment, Equation~(\ref{eqn:light-env-twls}), is then estimated for each image, yielding a triple of 9-D lighting environment coefficients, one per color channel. Because there was no substantive difference in the lighting environments across color channels, here we will report the lighting environments from a grayscale version of the input images.

Shown in Figure~\ref{fig:lighting-env-results} is a scatter plot of median ($50\%$ quantile) lighting environment estimates from the $50$ photographed (horizontal axis) and $50$ synthesized (vertical axis) images. To remove the impact of overall scene brightness, the lighting coefficients are divided by the magnitude of the zeroth-order term $Y_{0,0}(\cdot)$, leaving three first-order coefficients $Y_{1,*}(\cdot)$ and five second-order coefficients $Y_{2,*}(\cdot)$. The horizontal and vertical  bars correspond to the $35\%$ and $65\%$ quantile. For the most part, the synthesized and photographed lighting environments are well correlated, with a $R^2 = 0.80$ and with most of the coefficient medians lying near the dashed, zero-intercept, unit-slope line.

\begin{figure}[t]†
\begin{center} 
\begin{tabular}{c@{\hspace{0.15cm}}c@{\hspace{0.15cm}}c@{\hspace{0.15cm}}c}
     \includegraphics[width=0.225\linewidth]{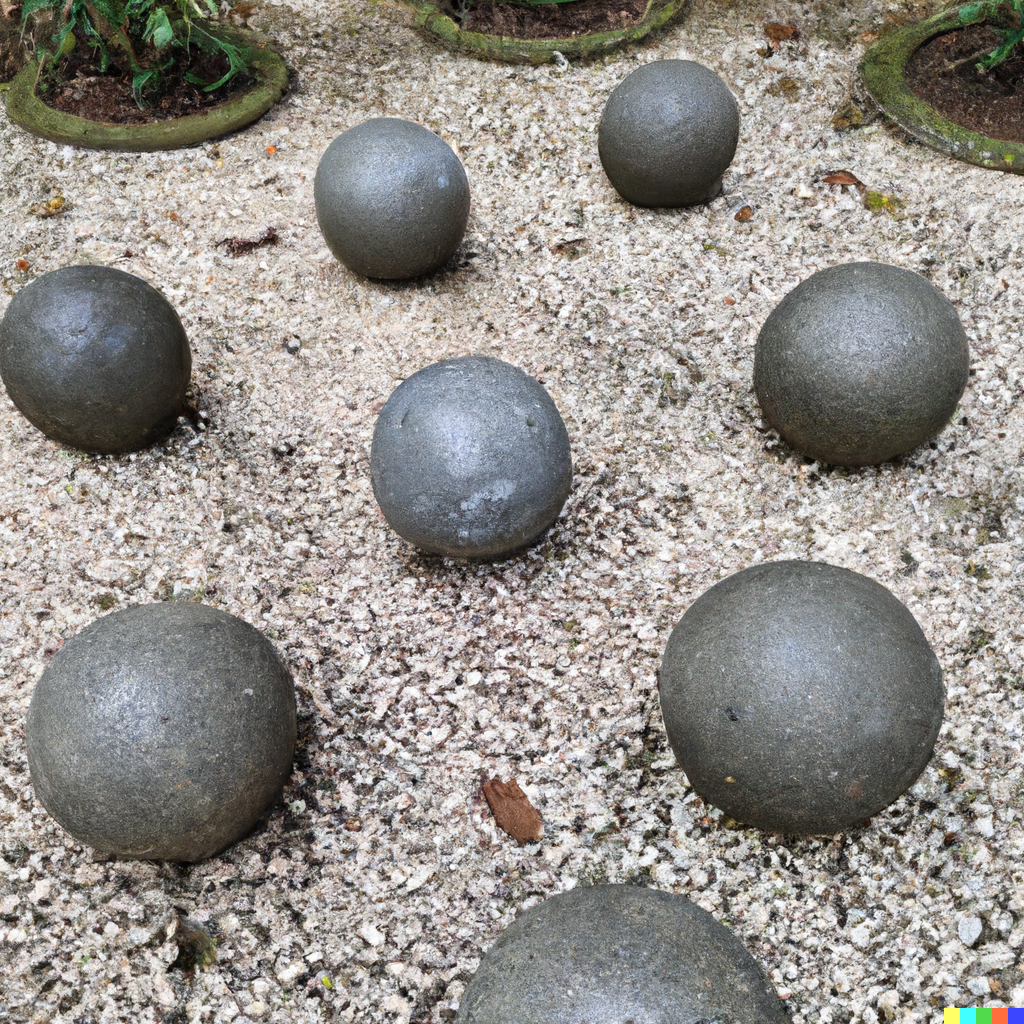} & 
     \includegraphics[width=0.225\linewidth]{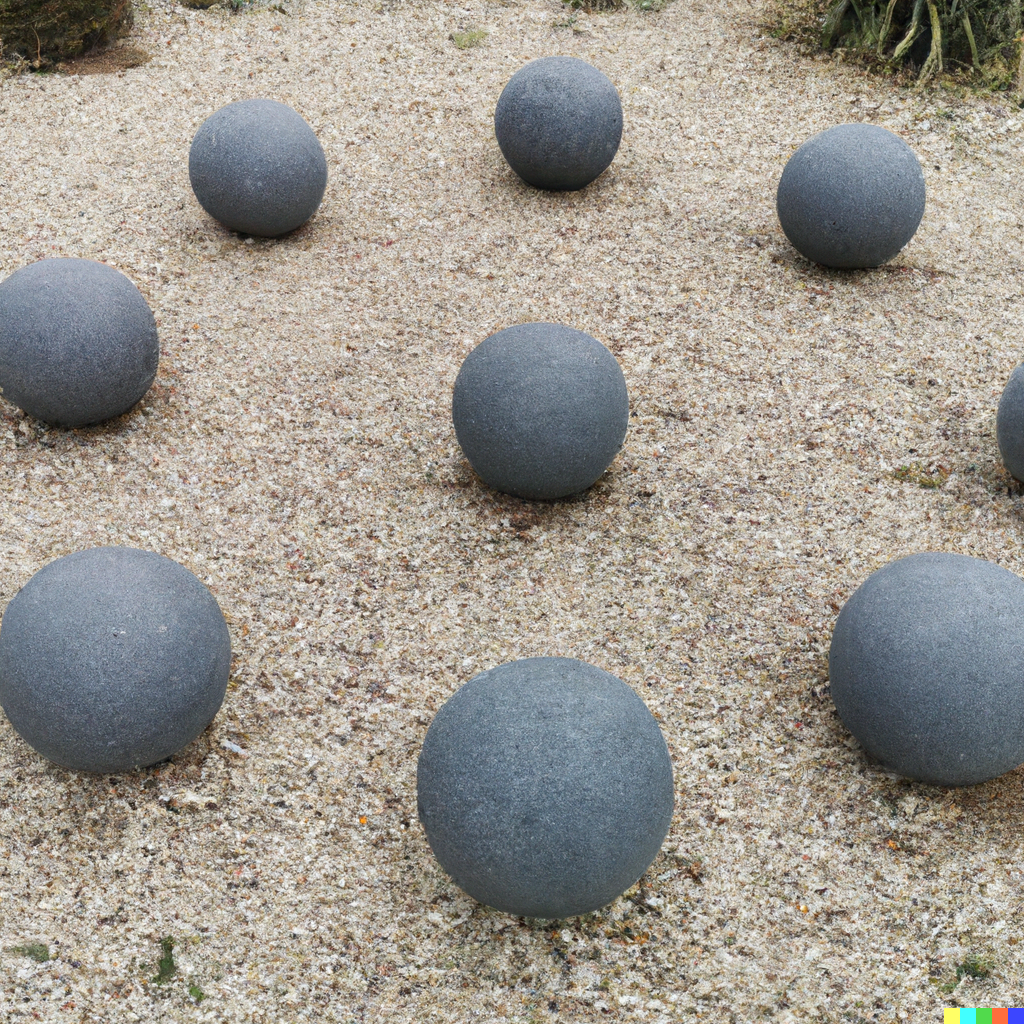} & 
     \includegraphics[width=0.225\linewidth]{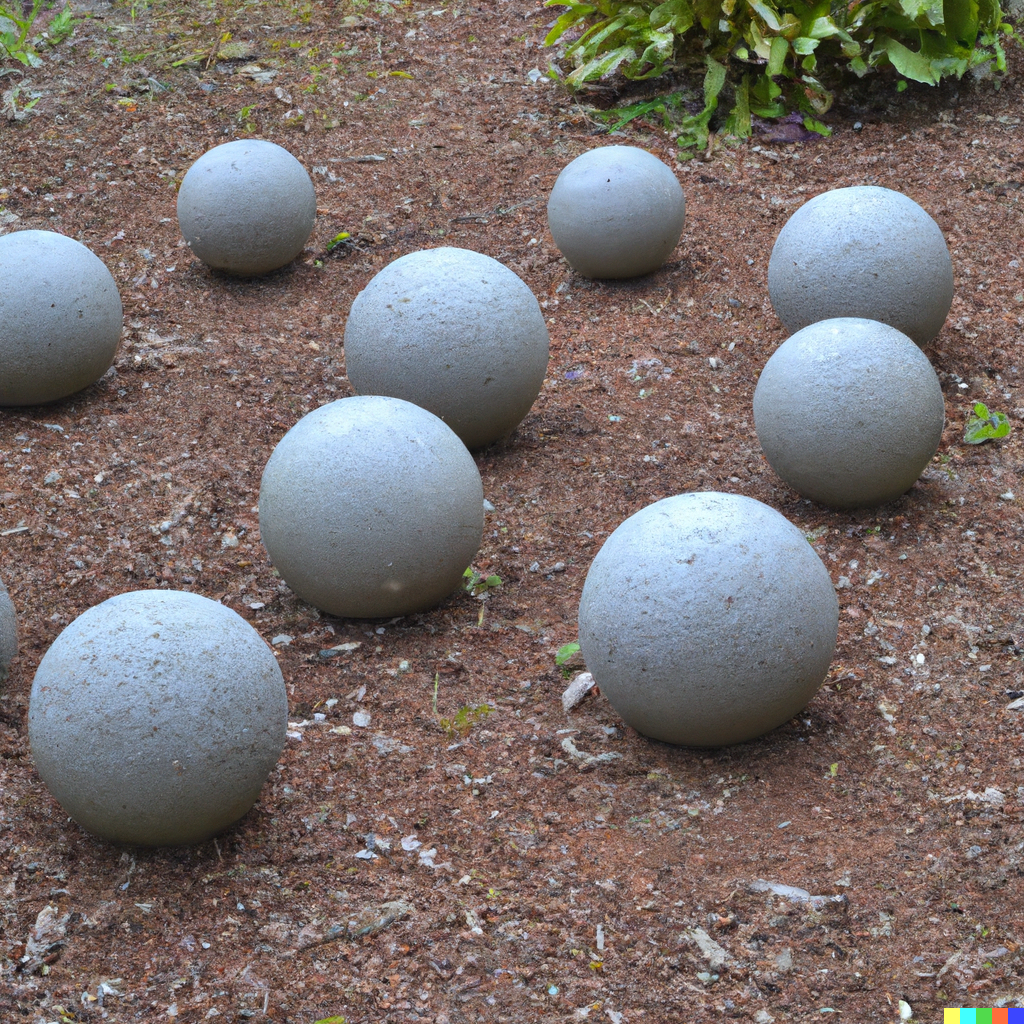} & 
     \includegraphics[width=0.225\linewidth]{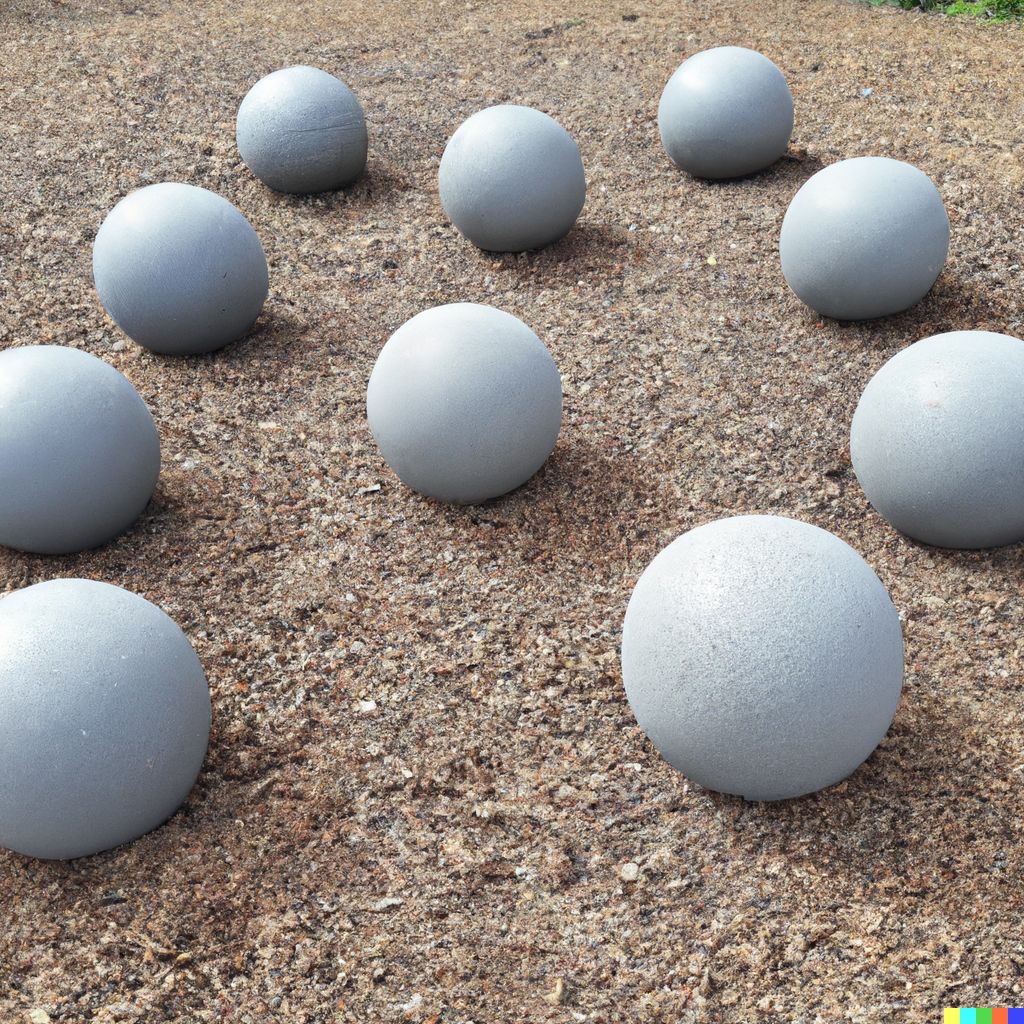}
\end{tabular}
\end{center}
\caption{Representative examples of synthesized images used to evaluate the consistency of lighting within a scene.}
\label{fig:spheres-5}
\end{figure}

Note that the coefficients on the $Y_{1,-1}$ spherical-harmonic term, corresponding to top/down illumination, is generally negative, corresponding to the typical natural illumination from above. The coefficients on the $Y_{1,0}$ term, corresponding to front/back illumination, is generally positive, corresponding to illumination behind the camera (an illumination in front of the camera places the subject in deep shadow or silhouette, and so photographers typically orient their subject so that they are illuminated with a light source behind the camera). 

The most notable deviation between synthesized and photographed lighting is the second-order term $Y_{2,0}(\cdot)$ which is significantly larger for synthesized images ($0.09$) than for photographed images ($-0.13$). As shown in Figure~\ref{fig:harmonics} (center, bottom row), this second-order spherical harmonic corresponds to illumination from in front of and behind the camera. This somewhat unnatural illumination pattern is prevalent in synthesized, but not in photographed images (the negative coefficient on this spherical harmonic corresponds to an intensity-inverted version of this pattern). Shown in Figure~\ref{fig:lighting-env-results} is a sphere rendered with the median photographic (top-left panel) and synthesized (bottom-left panel) lighting environment, revealing this front/back illumination difference (note these spheres are rendered with a different viewpoint than those in Figure~\ref{fig:harmonics} to highlight the differences between them).
  
For all lighting coefficients, there is significantly more variation in the environmental lighting for the synthesized images than photographed images, as seen by the difference in the vertical (simulated) and horizontal (photographed) bars. This larger variation motivates the next analysis in which we examine if the lighting within an image is consistent.

\begin{figure}[t]
\begin{center} 
\begin{tabular}{c@{\hspace{0.15cm}}c}
     \multicolumn{2}{c}{$Y_{0,0}$} \\
     \multicolumn{2}{c}{\includegraphics[width=0.475\linewidth]{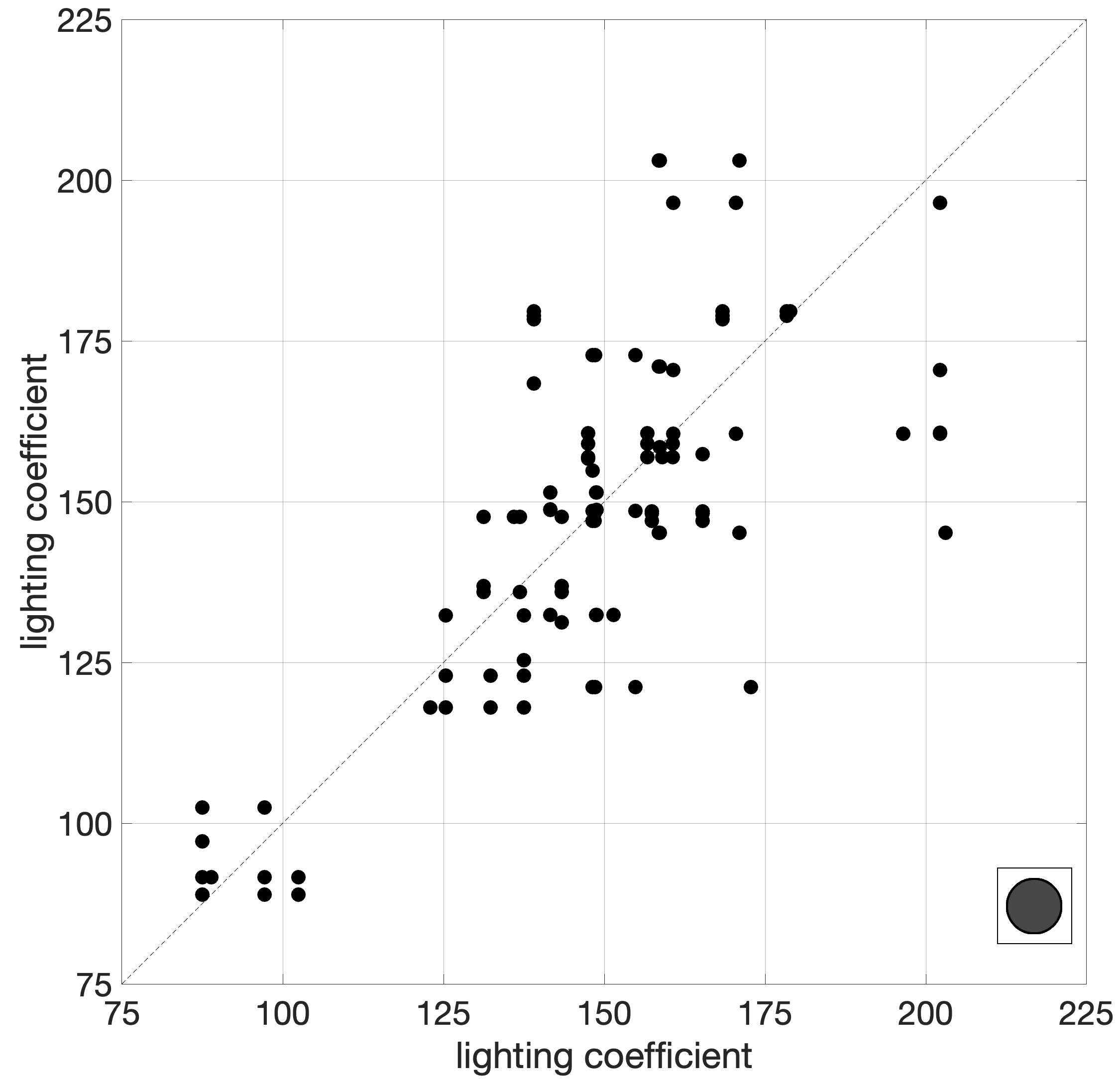}} \\
     $Y_{1,*}$ & $Y_{2,*}$ \\
     \includegraphics[width=0.475\linewidth]{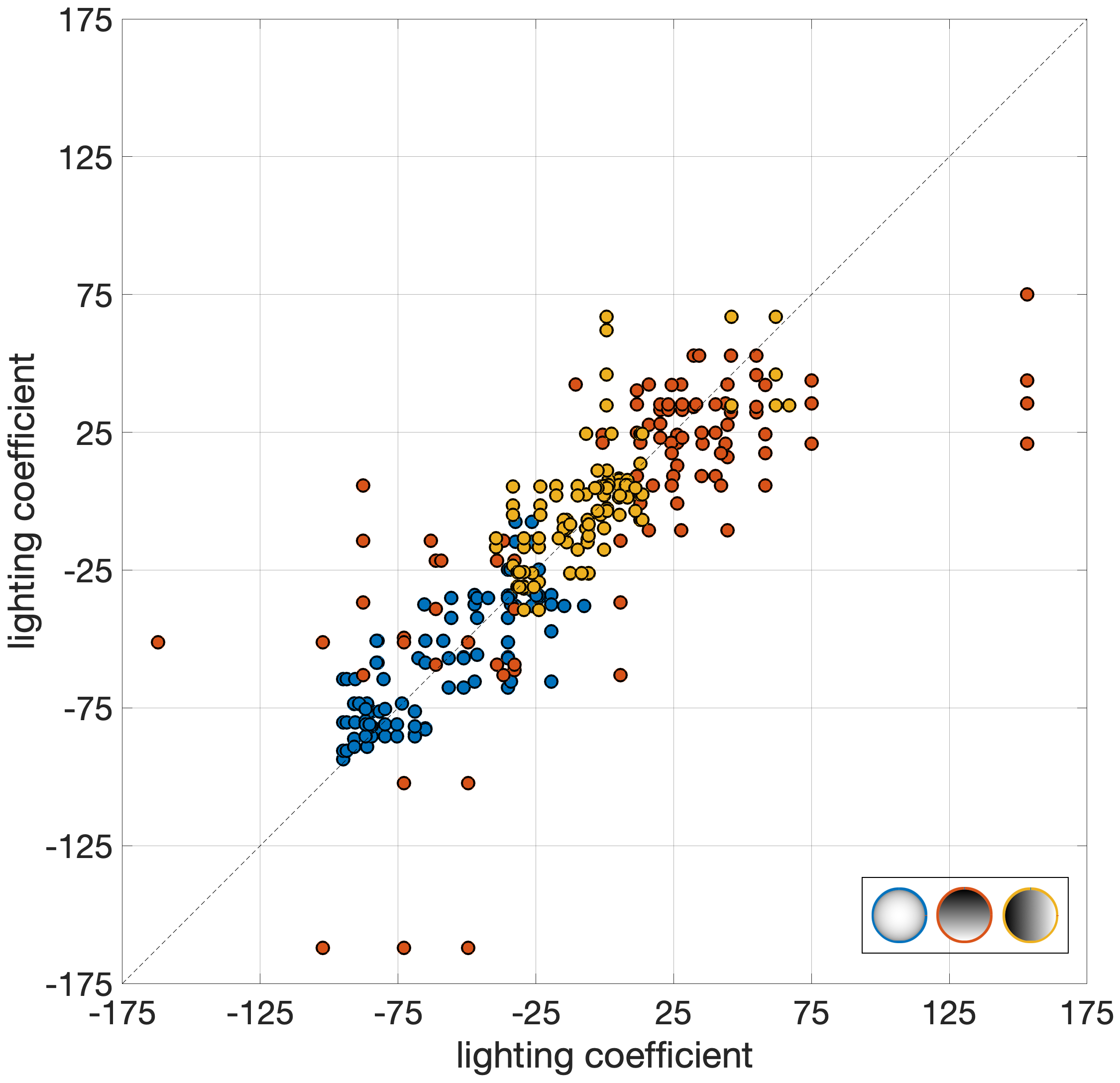} &
     \includegraphics[width=0.475\linewidth]{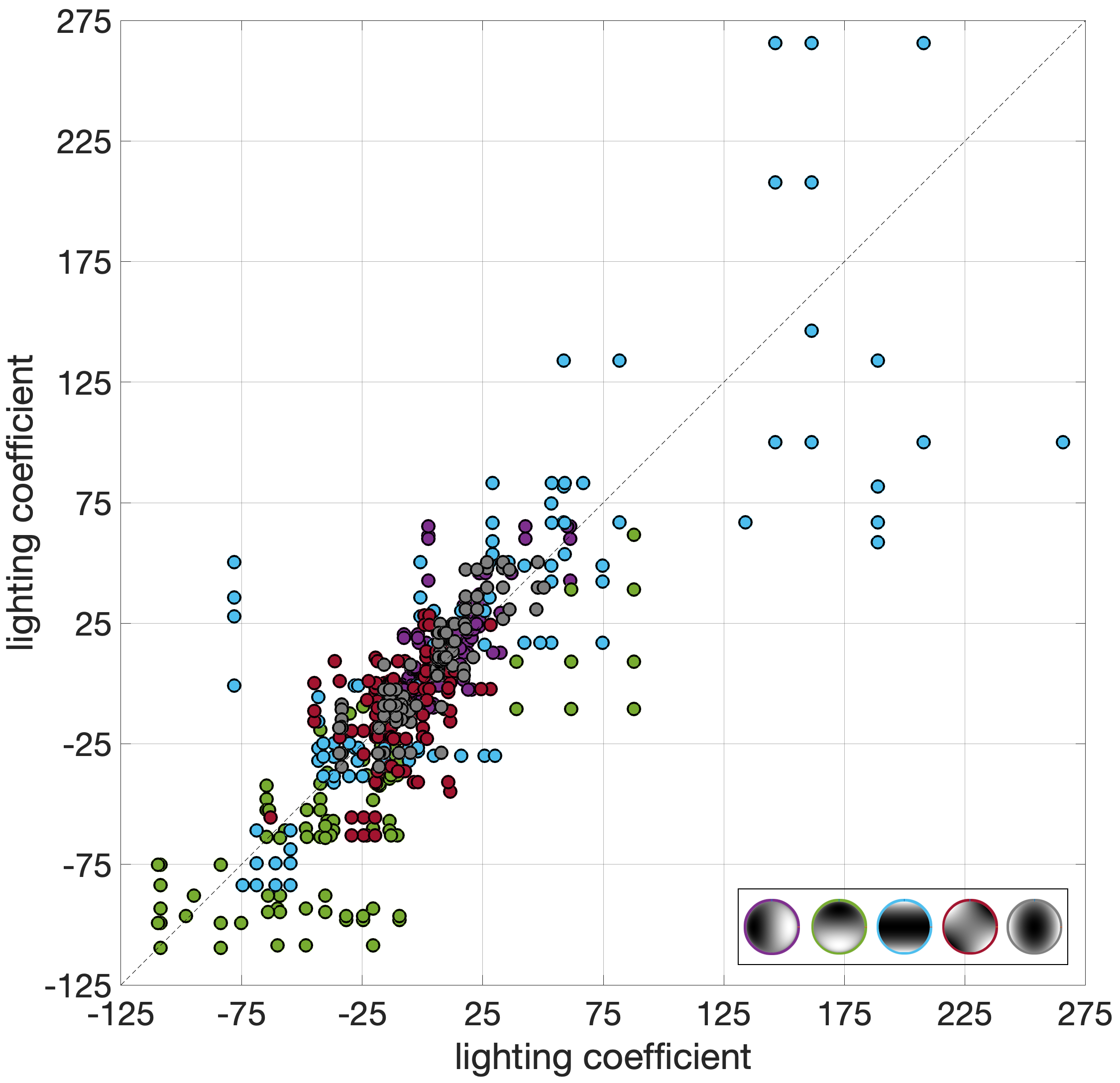}
\end{tabular}
\end{center}
\caption{The environmental lighting coefficients for all pairs of spheres in a single synthesized image. The correlation ($R^2$) of the zeroth- (top), first- (bottom left), and second-order (bottom right) spherical-harmonic coefficients is $0.54$, $0.69$, and $0.65$.}
\label{fig:lighting-env-results-5}
\end{figure}

A total of $10$ images were synthesized using \dalle with the text prompt ``a photo of a garden with gray matte spheres scattered on the ground.'' Shown in Figure~\ref{fig:spheres-5} are four representative examples of these synthesized images. From each image, the lighting environment was estimated from five unoccluded spheres. By comparing the lighting coefficients of all pairs of spheres, each image contributes $10$ lighting comparisons ($_{5}C_2 = (5 \times 4)/2 = 10$). Shown in Figure~\ref{fig:lighting-env-results-5} are three scatter plots corresponding to the correlation between the zeroth-, first-, and second-order spherical harmonics for these pair-wise comparisons within an image. Unlike the previous analysis, these lighting coefficients are not normalized because the scale of the coefficients should be the same within an image (assuming, per our model, distant light sources). With an $R^2$ of $0.54$, $0.69$, and $0.65$ for the zeroth-, first- and second-order harmonics, the within-image lighting consistency is weaker than the global across-image -- synthesized to photographed -- consistency ($R^2=0.80$), Figure~\ref{fig:lighting-env-results}. 

The lower correlation for the zeroth-order coefficients may be due to differences in the underlying  reflectance of the synthesized spheres. Although the first-order terms $Y_{1,-1}$ and $Y_{1,1}$ are fairly well correlated, the $Y_{1,0}$ term, corresponding to lighting from front/back, is significantly less correlated than the other first-order terms. Similarly, the second-order term $Y_{2,0}$ is significantly less correlated than the other second-order terms. These deviations are consistent with the previous analysis that found larger deviations in the lighting terms corresponding to these front/back lighting terms.

% =============================================================
\section{Discussion}
\label{sec:discussion}

Even in the absence of explicit 3-D modeling of a camera, scene geometry, and lighting -- as found in traditional CGI-rendering -- paint-by-text synthesis is capable of creating remarkably realistic images. We previously observed that specific perspective geometry is not always respected in these synthesized images~\cite{farid2022perspective}. Here we observe that globally, lighting is relatively consistent with natural photographed images, with a slight front/back (relative to the camera) lighting bias in synthesized images. We also observe that lighting within an image is somewhat consistent, but with some significant deviations. 

Because the images used in this study are fairly generic, there is no reason to believe these observations are specific to the relatively small number of analyzed images. It remains to be seen, however, if these observations will generalize to arbitrary indoor and outdoor scenes. This will require pairing photographed images with synthesized images from a broad range of different settings containing objects of known 3-D geometry.

Estimating 3-D lighting environments requires, of course, estimation of 3-D surface normals. While this is trivial for simple geometric shapes, it is not always easy or even possible to accurately estimate 3-D geometry from arbitrary objects in a single image. Three-dimensional modeling of faces~\cite{feng2021animatable} and bodies~\cite{bogo2016keep}, however, has become increasingly more reliable and accurate, allowing for the recovery of 3-D lighting environments from people in images. When 3-D models cannot be built, a 2-D lighting environment~\cite{johnson2007exposing} (or direction~\cite{nillius2001automatic,johnson2005exposing}) can be estimated by observing that five of the nine spherical harmonics -- $Y_{0,0}(\cdot)$, $Y_{1,-1}(\cdot)$, $Y_{1,1}(\cdot)$, $Y_{2,-2}(\cdot)$, and $Y_{2,2}(\cdot)$, Equation~(\ref{eqn:spherical-harmonics}) -- do not depend on the $z$-component of the required 3-D surface normal. As such, a 2-D, reduced dimensional, lighting environment can be estimated from only 2-D surface normals easily extracted from the boundary of arbitrary objects (where the $z$-component of the surface normal is $0$).

The trend in paint by text has been that increasing model size yields increasingly more realistic images. Given this trend, it remains to be seen if larger models will more accurately capture the naturalness and consistency in lighting environments. Until that time, however, physics-based forensic analyses should prove useful in analyzing this new breed of synthetic media.

% =============================================================
\section*{Acknowledgement}

Thanks to OpenAI for providing access to the \dalle synthesis engine.

% =============================================================
\bibliographystyle{unsrt}
\bibliography{refs}

% =============================================================
\end{document}